\documentclass[sigconf]{acmart}
\renewcommand\footnotetextcopyrightpermission[1]{} 
\usepackage{algorithm}
\usepackage{algorithmic}
\usepackage{subfigure}

\usepackage{flushend}

\usepackage[flushleft]{threeparttable}

\AtBeginDocument{%
  \providecommand\BibTeX{{%
    \normalfont B\kern-0.5em{\scshape i\kern-0.25em b}\kern-0.8em\TeX}}}

\begin{document}

\title{Non-Linear Fusion for Self-Paced Multi-View Clustering}

\author{Zongmo Huang}
\email{zongmohuang@gmail.com}
\affiliation{%
  \institution{University of Electronic Science and Technology of China}
  \city{Chengdu}
  \country{China}
}

\author{Yazhou Ren}
\authornote{Corresponding author.}
\email{yazhou.ren@uestc.edu.cn}
\affiliation{%
  \institution{University of Electronic Science and Technology of China}
  \city{Chengdu}
  \country{China}
}

\author{Xiaorong Pu}
\email{puxiaor@uestc.edu.cn }
\affiliation{%
  \institution{University of Electronic Science and Technology of China}
  \city{Chengdu}
  \country{China}
}

\author{Lifang He}
\email{lih319@lehigh.edu}
\affiliation{%
  \institution{Lehigh Univerisity}
  \city{Bethlehem}
  \state{PA}
  \country{USA}}

\renewcommand{\shortauthors}{Huang and Ren, et al.}

\begin{abstract}
  With the advance of the multi-media and multi-modal data, multi-view clustering (MVC) has drawn increasing attentions recently. 
  In this field, one of the most crucial challenges is that the characteristics and qualities of different views usually vary extensively.
  Therefore, it is essential for MVC methods to find an effective approach that handles the diversity of multiple views appropriately. 
  To this end, a series of MVC methods focusing on how to integrate the loss from each view have been proposed in the past few years.
  Among these methods, the mainstream idea is assigning weights to each view and then combining them linearly.
  In this paper, inspired by the effectiveness of non-linear combination in instance learning and the auto-weighted approaches, we propose Non-Linear Fusion for Self-Paced Multi-View Clustering (NSMVC), which is totally different from the the conventional linear-weighting algorithms.
  In NSMVC, we directly assign different exponents to different views according to their qualities.
  By this way, the negative impact from the corrupt views can be significantly reduced.
  Meanwhile, to address the non-convex issue of the MVC model, we further define a novel regularizer-free modality of Self-Paced Learning (SPL), which fits the proposed non-linear model perfectly.
  Experimental results on various real-world data sets demonstrate the effectiveness of the proposed method.
\end{abstract}



\maketitle

\section{Introduction}
As a fundamental field of machine learning, clustering \cite{hartigan1975clustering} has been studied extensively and a great number of classical clustering algorithms have been developed in the past few decades, such as $k$-means \cite{MacQueen:some}, density-based clustering \cite{ester1996density}, distribution-based clustering \cite{Banfield1993ModelbasedGA}, subspace-based clustering \cite{DRDSCJiang}, matrix factorization
based clustering \cite{NMFC}, hierarchical clustering \cite{Jain:clustering}, mean shift clustering \cite{Dorin2002mean}, and consensus clustering \cite{Strehl:cluster}.
Unfortunately, while in most real-world clustering tasks, an object can be usually described by multiple aspects, these conventional clustering methods only work on the single-view data.
To address this issue, a series of multi-view clustering (MVC) methods \cite{Kumar2011ACA,Kumar:2011:CMS:2986459.2986617,Tzortzis2012KernelBasedWM,Cai2013MultiViewKC,XU201625,zhang2017multi,KANG2020279,WANG20181,MMRNMFZong,PMSCXu, DIMCWen, WANG20191009,CGDTang} have been proposed recently and achieve much better clustering results comparing with their single-view counterparts. 


For MVC analysis, finding an appropriate approach to integrate different views is the foundation of making use of the complementary information within them. To tackle this problem, most existing MVC methods \cite{xu2016discriminatively, wang2014multi} are based on the following simple and intuitive idea: finding some measurements to weight each view and then combining them linearly, while the idea of non-linear fusion has been always neglected. 
On the other hand, the non-linear combining idea such as using $\ell_{2,1}$-norm has already been applied in instance learning \cite{FS21, Kong2011NMF, Cai2013MultiViewKC} and has shown better robustness comparing with the ordinary Frobenius norm.
As in most distance-based machine learning models, a few outliers with large losses always dominate the objective function and result in the poor performance of these algorithms.
By applying $\ell_{2,1}$-norm, the exponent of each sample's loss is decreased to 0.5 (i.e., rooted), thus the negative impact from the corrupt samples can be effectively alleviated and the robustness of the model can be enhanced.

Inspired by $\ell_{2,1}$-norm, there have been a few recent attempts of the implied weighting algorithms capable of mimicking its effects, with an aim of decreasing the exponent of the original loss function and thus achieving remarkable clustering results.
Enlighten by the idea of parameter-free learning, a series of MVC methods based on the auto-weighted (i.e. self-weighted) idea have been developed \cite{AMGLNie,SHI2020369,HSDDMVC,REN2020248}.
In these algorithms, instead of using some criterion to measure the quality of each view and then assigning weights to them, the weights of different views are directly generated from their loss values. 
Through this approach, the objective functions of these methods no longer have the form like the linear-weighted combination of the losses from each separate view.
However, in these models, the exponents of different views' losses are still the same, which means while smaller exponents alleviate the negative impact of the corrupt views, the influence of the reliable views are also weakened. 
Meanwhile, in the optimizing process, these methods still need to firstly transform their models into the traditional linear weighting forms. 

Different from the existing linear-weighting MVC methods, we propose non-linear fusion for self-paced multi-view clustering (NSMVC) to directly grant different exponents to different views based on their qualities. 
By this way, our method can alleviate the negative influence from the corrupt views in a non-linear manner, which is similar with the way of dealing with corrupt instances by $\ell_{2,1}$-norm.


Except for the challenge of effective view integration, most conventional MVC methods also face the non-convex problem and thus usually stuck into suboptimal local solutions.
To alleviate the non-convex issue, we further introduce the self-paced learning (SPL) mechanism \cite{kumar2010} to our model.
Imitating the learning process of human-beings, self-paced learning firstly trains the MVC model with the samples that have smaller losses, and as the iteration forwards, the samples with higher losses will gradually take part in the training process.
In this way, the noisy samples and ourliners will not join the training in the early iteration times, thus the robustness of MVC model can be significantly enhanced.
Meanwhile, to fit the non-linear MVC model, a novel regularizer-free self-paced learning modality is also developed in this work.
Through this SPL modality, the objective function of each view is completely constituted with the clustering losses of the selected instances and thus can avoid the influence from the value of the additional SPL regularizer.

Overall, we propose NSMVC to promote the clustering performance of MVC method in both view level and instance level.
In the view level, a novel learning paradigm based on non-linear fusion is developed to plenarily exploit the complementary information in different views. 
Moreover, to address the non-convex issue, we design a regularizer-free self-paced learning scenario to progressively train the MVC model from simplicity to complexity in the instance level.
Through this approach, the robustness of the MVC model can be significantly enhanced.

The main contributions of the paper are summarized as follows:
\begin{itemize}
\item To the best of our knowledge, this is the first attempt to develop a view-level non-linear fusion method in the multi-view clustering task.
\item A novel regularizer-free self-paced learning paradigm is designed to fit the non-linear model as well as alleviate the non-convex issue of conventional multi-view clustering method.
\item An effective optimizing approach to solve the proposed NSMVC model is derived, and experimental results on multiple real-world data sets demonstrate the effectiveness of the proposed method.
\end{itemize}

The rest of this paper is organized as follows. 
We give a brief review of the literature of multi-view clustering and self-paced learning in Section 2.
The details of the proposed DSMVC as well as its convergence and computational complexity analyses are presented in Section 3.
The experimental results and conclusion are respectively described in Sections 4 and 5.

\section{Related Work}
\subsection{Multi-View Clustering}
While in most real-world clustering tasks, objects can be described from multiple respects, the conventional clustering methods only can deal with single-view data.
To make full use of the complementary information from different views and obtain better clustering result, a great number of multi-view clustering methods have been proposed in the past decade. 
In co-training approach for multi-view spectral clustering (co-train) \cite{Kumar2011ACA} and co-regularized multi-view spectral clustering (co-reg) \cite{Kumar:2011:CMS:2986459.2986617}, Kumar et al. firstly put forward the fundamental assumption of MVC algorithms that among different views, the assignment of samples should be consistent.
Due to the diversity of the inherent characteristics of different views, it is important for MVC methods to find an appropriate integrating approach for each separate view.
To this end, a great number of relative MVC methods have been proposed in past few years.
Based on the kernel learning, Tzortzis and Likas \cite{Tzortzis2012KernelBasedWM} proposed multi-view kernel k-means clustering (MVKKM), in which each view is assigned with a weight according to its quality.
Xu et al \cite{XU201625} proposed the weighted multi-view clustering with feature selection (WMCFS), which weights different views based on their clustering performance and utilizes feature selection to promote the efficiency and effectiveness of the MVC model.
In multi-view clustering with multi-view capped-norm k-means (CAMVC) \cite{HUANG2018197}, Huang et al. assign weights for different views and implement the capped-norm loss in the objective function to achieve more stable clustering results with different initializations.

However, for most conventional linear-weighting methods, as they need some measurements to determine the weights of different views, it is inevitable to introduce additional hyper-parameters in their models.
According to the parameter-free principle, a series of auto-weighted MVC models have been proposed recently and have shown superior performance comparing with the conventional linear ones.

To decrease the number of parameters in MVC model, Nie et al. \cite{AMGLNie, 8047308} proposed an auto-weighted approach to assign weights to different views according to their losses.
Huang et al. \cite{HSDDMVC} further applied this idea into the deep matrix decomposition based MVC model and achieve remarkable results on various real-world data sets. 
In \cite{REN2020248}, Ren et al. not only use the auto-weighted approach to address the view quality issue, but also apply $\ell_{2,1}$-norm to tackle the noisy issue.
Since the weight of each view is generated by their loss, these models are actually represented in the non-linear forms and the negative impact from less reliable views is significantly alleviated.

Different from the existing methods that use the auto-weighting strategy for non-linear multi-view clustering, we directly assign different exponents to each view according to their qualities.

\subsection{Self-Paced Learning}
For most machine learning tasks, the non-convex issue is one of the most crucial factors that makes the models stuck into suboptimal solutions easily.
To tackle this issue, taking advantage of self-paced learning (SPL) \cite{kumar2010} and curriculum learning \cite{bengio2009Curriculum} can be an ideal solution.

Imitating the mechanism of human-learning, SPL firstly trains the model with easy samples and gradually select more complex samples in the following training steps.
In \cite{Jiang2015SPL}, Jiang et al. theoretically proved that applying SPL is beneficial in alleviating non-convex issue.
The general form of self-paced learning can be written as follows:
\begin{equation}\label{generalSPL}
    \begin{split}
        \min\limits_{v_i, w}\sum\limits_{i=1}^{n}&v_if(x_i,y_i,w)-\lambda\sum\limits_{i=1}^{n}v_i\\
        &s.t. \quad v_i \in \{0,1\}.
    \end{split}
\end{equation}

From Eq. (\ref{generalSPL}), only the samples with the loss smaller than $\lambda$ will take part in the training process.
As the iteration times increases, SPL gradually increases the value of $\lambda$ to let more samples join the training, thus the model will be trained from simplicity to complexity.

Due to the effectiveness of SPL, in these years, a series of MVC methods that utilizes SPL to promote the clustering performance have been proposed.
Xu et al. \cite{Xu2015Multi} were the first to use SPL in the multi-view clustering and reveal the applicability of SPL in solving MVC problem. 
In \cite{HZMDSMVC}, Huang et al. extended the idea of self-paced learning to feature learning and proposed a novel MVC method which alternatively performs sample learning and feature selection in a self-paced manner.

In this study, self-paced learning is not merely applied to tackle the non-convex issue, but also plays an important role in guiding the non-linear learning process.

\section{Non-Linear Fusion for Self-Paced Multi-View Clustering}
\subsection{Problem Definition}
Assuming that we are given a dataset with $n$ instances in $m$ views ${\{X^v\}}^m_{v=1}$, where $X^v = \{x^v_1,x^v_2,\ldots,x^v_n\} \in R^{d^v \times n}$, $d^v$ is the dimension of the feature vector in the $v$-th view.
Our target is to partition $n$ instances into $k$ clusters by making use of the complementary information from multiple views. 
Specifically, in this work, we aim to obtain better clustering results by utilizing the virtues of the non-linear fusion and self-paced learning.

\subsection{Proposed Model}
Inspired by the auto-weighted algorithms and non-linear learning's effectiveness in instance learning, in this paper, we develop a novel MVC model approach which is totally different from the conventional linear-combination form.
The model of our method can be written as:
\begin{equation}\label{eq:objor}
\begin{split}
    &\sum\limits^{m}_{v=1}{\phi(v)}^{\eta(v)}\\
    s.t.\quad \phi&(v) \geq 0,\quad\eta(v) \in (0,1],
\end{split}
\end{equation}
where $\phi(v)$ and $\eta(v)$ respectively represent the loss of the $v^{th}$ view and its exponent.

In Eq. (\ref{eq:objor}), the corrupt views are expected to be assigned with the smaller $\eta(v)$ values.
By this way, such views will be less influential during the optimizing process.
Considering the extreme circumstance, when the value of $\eta(v)$ approaches 0, the contribution of the $v^{th}$ view towards Eq. (\ref{eq:objor}) will be close to the constant 1, thus the $v^{th}$ view has no impact on the final clustering result.
By contrast, by granting more reliable views with higher $\eta(v)$ values, the MVC model will be more sensitive to the variations of the corresponded $\phi(v)$ values and these views will play more important roles during the training process.
Therefore, our proposed non-linear model in Eq. (\ref{eq:objor}) could significantly alleviate the negative influence from the corrupt views while maintain the availability of more reliable views.

To further address non-convex issue as well as fit the non-linear model shown in Eq. (\ref{eq:objor}), we design a novel regularizer-free SPL paradigm.
Concretely, we define $\phi(v)$ as:
\begin{equation}\label{phivloss}
    \phi(v)=\sum\limits^n_{i=1}\phi_{i}(v)=\sum\limits^n_{i=1}\lceil max(1-l^v_i/\lambda^v,0) \rceil \times l^v_i.
\end{equation}

\begin{figure}[!t]
  \centering
  \includegraphics[width= 0.75\linewidth]{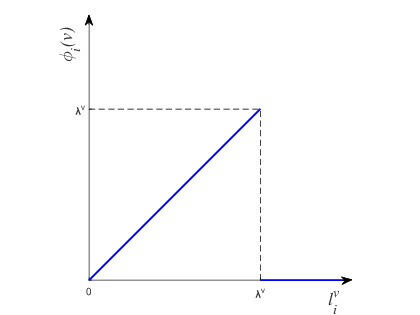}
  \caption{Illustration of the relation between $\phi_i(v)$ and $l^v_i$.}
  \label{spl}
\end{figure}

Here, $\phi_{i}(v) \geq 0$ denotes the $i^{th}$ sample's contribution to $\phi(v)$.
$\lambda^v \geq 0$ represents the control parameter of the self-paced learning in the $v^{th}$ view and $\lceil \cdot \rceil$ means the rounding up operation, while $l^v_i \geq 0$ and $n$ denote the loss of the $i^{th}$ instance in the $v^{th}$ view and the total number of samples respectively.
The relationship of $\phi_{i}(v)$ and $l^v_i$ can be described by Figure. \ref{spl}.
Specifically, the value of $l^v_i$ is computed by:

\begin{equation}\label{viloss}
    l^v_i={||x^v_i-C^vb_i||}_2^2.
\end{equation}

$C^v = \{c^v_1,c^v_2,\ldots,c^v_k\} \in R^{d^v \times k}$ represents the centers of clusters in the $v^{th}$ view, and $k$ denotes the predefined number of clusters. 
$B = \{b_1,b_2,\ldots,b_n\} \in R^{k \times n}$ reflects the clustering assignment of each data point and is shared by all the views. 
Concretely, if the $j^{th}$ sample is assigned to the $i^{th}$ cluster then $b_{ij}=1$, otherwise $b_{ij}=0$. 
Therefore, the value of $l^v_i$ physically means the squared Euclidean distance between the instance and the center of its belonging cluster. 


With Figure. \ref{spl}, Eq. (\ref{phivloss}) can be written in another form:
\begin{equation}\label{weightedMF}
    \phi(v)={||(X^v-C^vB)diag(w^v)||}_F^2,
\end{equation}
where $w^v=[w^v_1, w^v_2, \ldots, w^v_n]$, $w^v_i \in \{0, 1\}$, and $w^v_i=1$ only when $l^v_i \leq \lambda^v$.

From Eq. (\ref{weightedMF}), we can find out that Eq. (\ref{phivloss}) actually has the same selection result as the following formula in the conventional self-paced learning manner \cite{kumar2010} like Eq. (\ref{generalSPL}):
\begin{equation}\label{conventionalspl}
\begin{split}
    \min\limits_{w^v}&\sum\limits^{n}_{i=1} w^v_i l^v_i - w^v_i \lambda^v\\
    &s.t. \quad w^v_i \in \{0,1\}.
\end{split}
\end{equation}

When all the $\eta(v)$ values are equal to 1, solving Eq. (\ref{conventionalspl}) is equivalent to minimizing Eq. (\ref{weightedMF}). 
However, with the existance of an additional regularizer, the value of Eq. (\ref{conventionalspl}) is absolutely less than or equal to 0.
Due to this characteristic, such conventional SPL paradigm is not suitable for the non-linear model in Eq. (\ref{eq:objor}) that constrains $\eta(v) \in (0,1]$.
For instance, if $\eta(v)$ is 0.5, then the non-linear model works only when $\phi(v) \geq 0$, which is impracticable to the SPL paradigm in Eq. (\ref{conventionalspl}).
Instead, in Eq. (\ref{phivloss}), the novel SPL paradigm abandons the regularizer term.
By this way, the objective function is totally constituted with the clustering loss of the selected samples and thus ensures the non-negativity of $\phi(v)$.

During the iterative optimizing process, our method gradually increases the value of $\lambda^v$ until all the samples have joined the training, thus the MVC model will be trained from simplicity to complexity.

Another point that deserves our attention is that since $\lambda^v$ regulates the participation of samples in the $v^{th}$ view, itself can be an evidence of the quality of the corresponded view.
Generally speaking, a more reliable view has a smaller $\lambda(v)$ and vice versa.
With the constraint that $\eta(v) \in (0,1]$, the value of $\eta(v)$ is:
\begin{equation}\label{update:eta}
    \eta(v)=\frac{\min\limits_{u}{\lambda^u}}{\lambda^v}.
\end{equation}

Integrating Eq. (\ref{eq:objor}) to (\ref{update:eta}), the objective of NSMVC becomes:
\begin{equation}\label{eq:obj}
    \min\limits_{C^v, B, w^v}\sum\limits^{m}_{v=1}{{||(X^v-C^vB)diag(w^v)||}_F^{2{\min\limits_{u}{\lambda^u}}/{\lambda^v}}}.
\end{equation}

\subsection{Optimization}
To start the optimizing process, our method firstly initializes the cluster centers $C^v$ and assignment matrix $B$ randomly.
After that, the objective function Eq. (\ref{eq:obj}) is optimized by iteratively updating each variable while others are fixed.

\subsubsection{\texorpdfstring{Step1: Fix $C^v$, $B$, update $\lambda^v$, $w^v$, and $\eta(v)$}{pdfstring}}\label{alterntivelyupdating}
\ 

To take the advantage of self-paced learning and enhance the robustness of the MVC model, our method should firstly determine the participation of samples in the beginning of each iteration.
Assuming that the whole SPL process needs $T$ iterations, the updating rule of $\lambda^v$ in the $t^{th}$ iteration is:
\begin{equation}\label{eq:update(lambda)}
    \lambda^v=min(l^v_i)+(\alpha+(t-1)\times\beta)\times(max(l^v_i)-min(l^v_i))
\end{equation}
In Eq. (\ref{eq:update(lambda)}), the value of $\beta$ is computed by:
\begin{equation}\label{beta}
\begin{split}
    &\beta=\frac{1-\alpha}{T-1}\\
    s.&t. \quad \alpha \in [0,1].
\end{split}
\end{equation}
By this way, we could control the starting point of the self-paced learning process as well as finish it in the promised iteration times $T$.
For instance, when we set the start point $\alpha$ as 0.5 and $T$ as 6, from Eq. (\ref{eq:update(lambda)}), the value of $\beta$ will become 0.1. 
Then, in the first iteration, only the samples with losses smaller than the mean value of the $max(l^v_i)$ and the $min(l^v_i)$ will be selected.
In the final ($6^{th}$) iteration, the value of $\lambda^v$ will be just large enough to let all the samples join the training.

After deciding the value of $\lambda^v$ for $v^{th}$ view, the values in $w^v$ can be obtained naturally. 
When all $\lambda^v$ are acquired, the value of each $\eta(v)$ will be calculated by Eq. (\ref{update:eta}).

\subsubsection{\texorpdfstring{Step2: Fix $\lambda^v$, $\eta(v)$, and $w^v$, update $C^v$ and $B$ alternately}{pdfstring}}
\ 

\textbf{(a) Fix $B$, update $C^v$: }

When $\eta(v)$, $w^v$, and $B$ are fixed, optimizing Eq. (\ref{eq:obj}) is equivalent to solving the following problem for each view:
\begin{equation}\label{eq:objC}
    \min\limits_{C^v}\sum\limits^{m}_{v=1}{{||(X^v-C^vB)diag(w^v)||}_F^2}^{\eta(v)}.
\end{equation}

From Eq. (\ref{eq:objC}), the optimal $C^v$ of each view can be separately obtained by finding the solution of:
\begin{equation}\label{eq:objCF}
    \min\limits_{C^v}{{||(X^v-C^vB)diag(w^v)||}_F^2}.
\end{equation}

Then, Eq. (\ref{eq:objCF}) can be transformed into the form of matrix's trace:
\begin{equation}\label{eq:objCTr}
    \min\limits_{C^v}Tr\big((X^v-C^vB)diag^2(w^v){(X^v-C^vB)}^T\big).
\end{equation}

Regarding Eq. (\ref{eq:objCTr}) as a function $J(C^v)$ and its gradient is:
\begin{equation}
\frac{\partial J}{\partial C^v}=2X^vdiag^2(w^v)B^T\\
-2C^vBdiag^2(w^v)B^T.
\end{equation}
Setting this gradient to \textbf{0}, then the updating rule of $C^v$ is:
\begin{equation}\label{update:C}
    C^v=X^vdiag^2(w^v)B^T{(Bdiag^2(w^v)B^T)}^{-1}.
\end{equation}

\begin{algorithm}[!t] \small
\renewcommand{\algorithmicrequire}{\textbf{Input:}}
\renewcommand{\algorithmicensure}{\textbf{Output:}}
\caption{The NSMVC Algorithm.}
\label{alg:algorithm}
\begin{algorithmic}[1]
\REQUIRE ~~
Data set $X^v$, $v=1,2,\ldots,m$; Cluster number $k$; SPL start point $\alpha$ and iteration times $T$.
\ENSURE 
The final cluster center matrix $C^v$, assignment matrix $B$, $v=1,2,\ldots,m$. 
\STATE Initialize $C^v$ and $B$ randomly.
\REPEAT 
\FOR{each view $v$}
\STATE According to Eq. (\ref{eq:update(lambda)}) update $\lambda^v$ to let more samples join the training.
\FOR{each sample $i$}
\STATE Update $w^v_i=1$ if $l^v_i<=\lambda^v$, otherwise $w^v_i=0$.
\ENDFOR
\ENDFOR
\STATE Update $\eta(v)$ for each view according to Eq. (\ref{update:eta}).
\REPEAT
\FOR{each view $v$}
\STATE Fix $\eta(v), w^v$ and $B$, update $C^v$ according to Eq. (\ref{update:C}).
\ENDFOR
\STATE Fix $\eta(v), C^v$ and $w^v$, update $B$ according to Eqs. (\ref{thetav}) and (\ref{update:B}).
\UNTIL{convergence or exceed the maximal number of iterations}
\UNTIL{all data points are selected}
\STATE \textbf{return} $C^v$ and $B$, $v=1,2,\ldots,m$.
\end{algorithmic}
\end{algorithm}

\textbf{(b) Fix $C^v$, update $B$: }

As for assignment matrix $B$, due to the non-linearity of our model, it cannot be solved by the conventional route that divides the loss function in the instance level and finds the optimal $b_i$ for each sample separately.
To address this issue, we design a simple and effective solving approach, which guarantees to decrease the value of objective function with the time complexity $O(n)$. 

Concretely, we sequentially update $b_i$ for each sample one by one. 
When updating $b_i$, we define $\theta_{i}(v)$ for the $i^{th}$ sample in the $v^{th}$ view as:
\begin{equation}\label{thetav}
    \theta_{i}(v)= \phi(v)-l^v_i.
\end{equation}
Substituting Eq. (\ref{thetav}) in Eq. (\ref{eq:objC}), $b_i$ can be obtained by solving:
\begin{equation}\label{update:B}
    \mathop{\arg\min}_{b_i}\sum\limits^{m}_{v=1}{(\theta_{i}(v)+||(x^v_i-C^vb_i)w^v_i||_2^2)}^{\eta(v)}.
\end{equation}
Since the possible alternatives' number of $b_i$ is the predefined cluster number $k$, Eq. (\ref{update:B}) can be easily addressed by exhaustive search.
When the optimal $b_i$ is obtained according to Eq. (\ref{update:B}), the $\phi(v)$ value in Eq. (\ref{thetav}) for each view can be updated and will be used to find the optimal $b_{(i+1)}$ for the ${(i+1)}^{th}$ sample. 
As we usually have the fact that $k << n$, this step only needs $O(n)$ operations.

\begin{table*}[t] 
  \normalsize
\begin{threeparttable}
  \centering
 \caption{Summary of the data sets used in the experiments.}
    \begin{tabular}{ccccccl}
    \hline View  & Handwritten Numerals & MSRCv1 & Cornell & Texas & Washington & Wisconsin \\
    \hline
    1 & Profile correlations (216) & Color
Moments (24) & Citation (195) & Citation (187) & Citation (230) & Citation (265) \\
    2 & Fourier coefficients (76) & HOG (576) & Content (1703) & Content (1398) & Content (2000) & Content (1703) \\
    3 & Karhunen coefficients (64) & GIST (512) & - & - & - & - \\
    4 & Morphological (6) & LBP (256) & - & - & - & - \\
    5 & Pixel averages (240) & Centrist (254) & - & - & - & - \\
    6 & Zernike moments (47) & - & - & - & - & - \\
    \# Samples & 2000 & 210 & 195 & 187 & 230 & 265 \\
    \# Classes & 10 & 7 & 5 & 5 & 5 & 5 \\
    \hline
    \end{tabular}%
    \label{1}%
\begin{tablenotes}
  \item[*] Numbers in parentheses are the number of features in each view.
  \end{tablenotes}
\end{threeparttable}
\end{table*}%

\begin{figure*}[!t]
\centering
 \subfigure[Handwritten]{\label{fig:Handwritten}
    \begin{minipage}[l]{0.4\linewidth}
     \centering
     \includegraphics[width=2.5in]{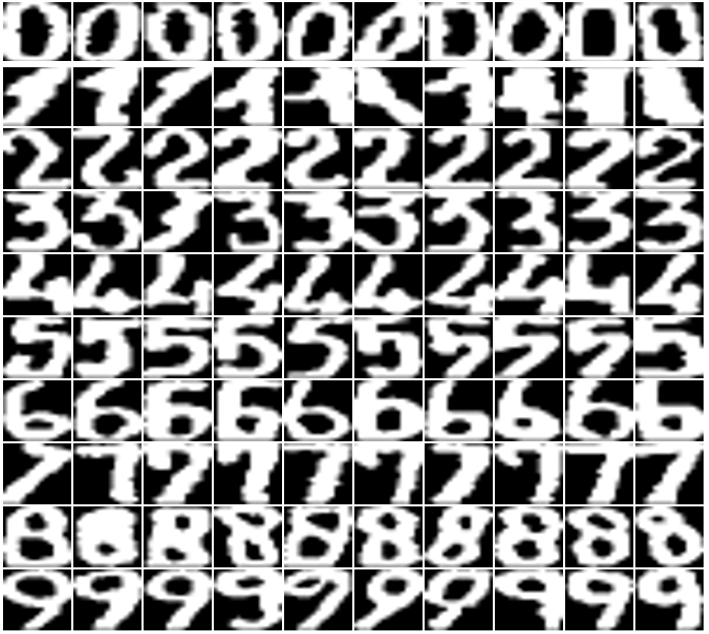}
    \end{minipage}
 }
 \subfigure[MSRCv1]{\label{fig:MSRCv1}
    \begin{minipage}[l]{0.4\linewidth}
     \centering
     \includegraphics[width=2.5in]{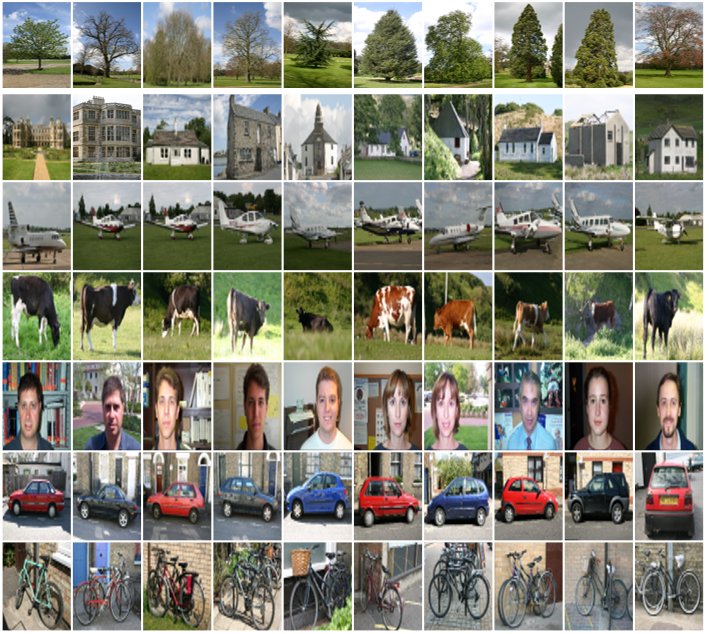}
    \end{minipage}
 }
\caption{An example of two image data sets. Each row represents ten image samples from a single class.}
\label{imgsamples}
\end{figure*}

In \textit{Step2}, $C^v$ and $B$ are alternatively updated until convergence or exceed the maximal iteration times.

The above three steps correspond to an entire iteration of the optimization. 
Our method keeps the optimization process running until all the samples have taken part in the training process. When the whole training process is finish, the final cluster center matrices $C^1,C^2,\ldots,C^m$ and assignment matrix $B$ reflect the clustering result.

The entire procedure of NSMVC is summarized in Algorithm \ref{alg:algorithm}.




\subsection{Convergence Analysis}
\label{sec:convergence_analysis}
As the NSMVC finishes in a fixed number of iterations and \textit{Step 1} only plays an instance selection role for the next step, thus we only need to focus on the convergence trend in the alternatively updating process of \textit{Step 2}.
In \textit{Part (a)} of \textit{Step 2}, $C^v$ is updated by finding the optimal solution of Eq. (\ref{eq:objC}), so that the value of Eq. (\ref{eq:obj}) is guaranteed to decrease.
In \textit{Part (b)} of \textit{Step 2}, as we find the optimal $b_i$ for each instance sequentially, the value of Eq. (\ref{eq:obj}) decreases monotonously.
With the non-negativity of $\phi(v)$ and the monotone bounded convergence theorem, NSMVC is guaranteed to converge to a local minimum.

\subsection{Computational Complexity Analysis}
Let $D$ and $P$ denote the maximal feature dimensionality and the maximal number of iterations for alternatively updating $C^v$ and $B$.
For \textit{Step 1}, as the $\lambda^v$ of each view is generated the from the minimal and maximal clustering loss of all the samples, obtaining $\lambda^v$ needs $O(nD)$ operations.
Thus, updating all the $\lambda^v (v=1,2,\dots,m)$ needs $O(nmD)$ operations.
After that, the values of $w^v$ and $\eta(v)$ can be naturally computed with $O(n)$ and $m$ operations respectively.
Therefore, the time complexity of \textit{Step 1} is $O(nD)$.
For \textit{Step 2}, since $diag(w^v)$ is diagonal, the time complexity of updating $C^v$ by Eq. (\ref{update:C}) is $O(nDk)$. 
Therefore, updating all the $C^v (v=1,2,\dots,m)$ needs $O(nmDk)$ operations.
Then, as discussed in the \textit{Part (b)} of \ref{alterntivelyupdating}, updating $B$ needs $O(n)$ operations,  thus the total time complexity of \textit{Step 2} is $O(PnmDk)$.
Since the entire self-paced learning process needs $T$ iteration times, the overall computational complexity of NSMVC is $O(PTnmDk)$.
With the fact that $T\ll n$ usually holds and applying $k$-means individually on $m$ views needs $O(PmnDk)$ operations.
In summary, the proposed NSMVC shares similar computational complexity with the traditional $k$-means method, which is linear to the data size $n$.

\begin{table}[!t] 
  \normalsize
  \centering
  \caption{Results on Handwritten Numerals.}
     \setlength{\tabcolsep}{3mm}{\begin{tabular}{cccl}
    \hline Methods  & ACC(\%) & Purity(\%) & NMI(\%) \\
    \hline
    KM(1) & 57.71$\pm$4.98 & 63.71$\pm$3.69 & 60.27$\pm$2.41  \\
    KM(2) & 62.99$\pm$6.72 & 65.38$\pm$5.03 & 64.38$\pm$2.76  \\
    KM(3) & 70.53$\pm$7.27 & 73.38$\pm$5.84 & 70.93$\pm$3.59 \\
    KM(4) & 38.09$\pm$1.55 & 43.80$\pm$0.97 & 47.76$\pm$0.24 \\
    KM(5) & 70.05$\pm$6.89 & 72.56$\pm$6.36 & 70.38$\pm$3.99 \\
    KM(6) & 52.10$\pm$2.93 & 55.95$\pm$2.33 & 50.01$\pm$1.82 \\
    KM(All) & 50.72$\pm$4.17 & 56.01$\pm$2.44 & 57.37$\pm$1.64 \\
    Co-train & 73.28$\pm$5.87 & 74.92$\pm$3.89 & 71.04$\pm$2.15 \\
    Co-reg & 78.09$\pm$6.89 & 80.63$\pm$5.36 & 75.50$\pm$2.91 \\
    MVKKM & 62.18$\pm$3.34 & 65.56$\pm$2.40 & 65.80$\pm$1.19  \\
    AMGL & 81.22$\pm$6.53 & 84.24$\pm$4.99 & 86.89$\pm$2.66 \\
    CAMVC & 74.98$\pm$7.96 & 78.84$\pm$6.90 & 78.07$\pm$4.25 \\
    MSPL & 80.26$\pm$3.93 & 83.60$\pm$3.22 & 82.80$\pm$2.25 \\
    SAMVC & 75.37$\pm$12.71 & 79.74$\pm$11.91 & 82.62$\pm$12.49 \\
    DMVC  & 79.91$\pm$8.56 & 83.77$\pm$6.72 & 85.18$\pm$4.01 \\
    NSMVC & \textbf{88.52$\pm$6.40} & \textbf{90.53$\pm$4.63} & \textbf{89.10$\pm$2.25}\\
    \hline
    \end{tabular}}%
  \label{2}%
\end{table}%

\begin{table}[!t]  
  \normalsize
  \centering
  \caption{Results on MSRCv1.}
     \setlength{\tabcolsep}{3mm}{\begin{tabular}{cccl}
    \hline Methods  & ACC(\%) & Purity(\%) & NMI(\%) \\
    \hline
    KM(1) & 35.76$\pm$2.38 & 37.88$\pm$2.45 & 24.25$\pm$2.50  \\
    KM(2) & 62.69$\pm$6.60 & 64.60$\pm$5.59 & 54.16$\pm$4.45  \\
    KM(3) & 62.00$\pm$5.52 & 64.90$\pm$4.17 & 57.03$\pm$3.74 \\
    KM(4) & 47.29$\pm$1.55 & 49.55$\pm$0.97 & 41.38$\pm$0.24 \\
    KM(5) & 54.64$\pm$6.79 & 55.55$\pm$5.25 & 45.18$\pm$3.19 \\
    KM(All) & 46.29$\pm$3.10 & 46.29$\pm$3.03 & 42.07$\pm$2.18 \\
    Co-train & 66.55$\pm$5.77 & 69.33$\pm$4.46 & 58.18$\pm$3.46 \\
    Co-reg & 41.52$\pm$4.31 & 44.21$\pm$3.89 & 35.36$\pm$3.61 \\
    MVKKM & 70.19$\pm$3.73 & 70.95$\pm$3.31 & 61.61$\pm$3.07  \\
    AMGL & 69.74$\pm$7.04 & 71.81$\pm$5.01 & \textbf{68.35$\pm$3.24} \\
    CAMVC & 67.88$\pm$5.18 & 71.14$\pm$3.49 & 62.86$\pm$2.73 \\
    MSPL & 50.19$\pm$6.01 & 52.29$\pm$4.94 & 43.30$\pm$2.66 \\
    SAMVC & 65.31$\pm$8.82 & 68.19$\pm$8.01 & 61.57$\pm$6.28 \\
    DMVC  & 62.57$\pm$10.49 & 63.67$\pm$10.48 & 58.75$\pm$8.67 \\
    NSMVC & \textbf{74.65$\pm$5.62} & \textbf{77.14$\pm$4.44} & \textbf{66.65$\pm$5.00}\\
    \hline
    \end{tabular}}%
  \label{3}%
\end{table}%

\begin{table}[!t]  
  \normalsize
  \centering
  \caption{Results on Cornell.}
     \setlength{\tabcolsep}{3mm}{\begin{tabular}{cccl}
    \hline Methods  & ACC(\%) & Purity(\%) & NMI(\%) \\
    \hline
    KM(1) & 42.70$\pm$2.14 & 44.96$\pm$1.00 & 8.60$\pm$3.36  \\
    KM(2) & 45.56$\pm$5.87 & 48.56$\pm$3.58 & 12.34$\pm$5.42  \\
    KM(All) & 47.47$\pm$6.42 & 49.69$\pm$5.01 & 13.54$\pm$6.96 \\
    Co-train & 40.62$\pm$1.27 & 46.41$\pm$0.88 & 14.48$\pm$1.40 \\
    Co-reg & 42.39$\pm$1.09 & 44.10$\pm$0.36 & 5.65$\pm$2.45 \\
    MVKKM & 41.64$\pm$3.72 & 44.72$\pm$1.03 & 7.27$\pm$1.83  \\
    AMGL & 42.68$\pm$0.40 & 43.81$\pm$0.26 & 3.74$\pm$0.37 \\
    CAMVC & 44.10$\pm$2.74 & 49.16$\pm$2.12 & 9.81$\pm$4.95 \\
    MSPL & 44.09$\pm$3.18 & 46.19$\pm$2.65 & 8.78$\pm$4.62 \\
    SAMVC & 43.43$\pm$0.82 & 44.69$\pm$0.50 & 6.82$\pm$3.11 \\
    DMVC  & 44.30$\pm$2.74 & 46.72$\pm$2.82 & 12.42$\pm$3.99 \\
    NSMVC & \textbf{50.62$\pm$6.45} & \textbf{60.21$\pm$2.76} & \textbf{26.56$\pm$3.94} \\
    \hline
    \end{tabular}}%
  \label{4}%
\end{table}%
  
  \begin{table}[!t]  
  \normalsize
  \centering
  \caption{Results on Texas.}
     \setlength{\tabcolsep}{3mm}{\begin{tabular}{cccl}
    \hline Methods  & ACC(\%) & Purity(\%) & NMI(\%) \\
    \hline
    KM(1) & 55.51$\pm$1.63 & 57.18$\pm$1.25 & 7.90$\pm$4.50  \\
    KM(2) & 55.56$\pm$6.07 & 60.04$\pm$4.61 & 16.18$\pm$10.90  \\
    KM(All) & \textbf{56.63$\pm$5.84} & 60.41$\pm$5.04 & 14.63$\pm$10.38 \\
    Co-train & 48.13$\pm$2.75 & 58.00$\pm$0.89 & 14.22$\pm$1.78 \\
    Co-reg & 53.37$\pm$2.67 & 56.04$\pm$0.22 & 4.57$\pm$1.89 \\
    MVKKM & 52.84$\pm$5.63 & 56.84$\pm$0.89 & 7.70$\pm$3.71  \\
    AMGL & 56.13$\pm$0.43 & 56.84$\pm$0.25 & 5.43$\pm$0.51 \\
    CAMVC & \textbf{59.23$\pm$4.49} & 60.71$\pm$4.42 & 14.92$\pm$9.54 \\
    MSPL & \textbf{56.68$\pm$3.85} & 58.97$\pm$2.94 & 11.30$\pm$6.30 \\
    SAMVC & 56.81$\pm$1.38 & 57.68$\pm$1.35 & 8.54$\pm$4.54 \\
    DMVC  & \textbf{56.90$\pm$4.29} & 59.84$\pm$3.24 & 16.32$\pm$7.15 \\
    NSMVC & \textbf{57.81$\pm$4.93} & \textbf{67.17$\pm$1.45} & \textbf{25.23$\pm$2.60} \\ 
    \hline
    \end{tabular}}%
    \label{5}%
\end{table}%

\begin{table}[!thb]  
  \normalsize
  \centering
  \caption{Results on Washington.}
     \setlength{\tabcolsep}{3mm}{\begin{tabular}{cccl}
    \hline Methods  & ACC(\%) & Purity(\%) & NMI(\%) \\
    \hline
    KM(1) & 49.80$\pm$6.84 &51.46$\pm$6.86 & 8.44$\pm$7.84  \\
    KM(2) & \textbf{57.54$\pm$9.77} & 61.39$\pm$9.64 & 25.36$\pm$14.50  \\
    KM(All) & \textbf{58.75$\pm$9.40} & 66.29$\pm$8.68 & 26.23$\pm$12.89 \\
    Co-train & 53.99$\pm$2.25 & 62.93$\pm$1.22 & 19.30$\pm$1.78 \\
    Co-reg & 55.97$\pm$2.95 & 58.64$\pm$4.19 & 16.68$\pm$3.63 \\
    MVKKM & 48.39$\pm$1.85 & 49.49$\pm$1.83 & 8.65$\pm$4.53  \\
    AMGL & 47.26$\pm$0.20 & 48.26$\pm$0.00 & 3.58$\pm$0.32 \\
    CAMVC & \textbf{58.97$\pm$10.57} & 60.81$\pm$10.62 & 22.53$\pm$14.99 \\
    MSPL & 52.67$\pm$7.99 & 54.16$\pm$7.95 & 13.75$\pm$11.15 \\
    SAMVC & 52.93$\pm$7.95 & 53.70$\pm$8.12 & 11.39$\pm$9.43 \\
    DMVC  & \textbf{58.45$\pm$6.96} & 62.44$\pm$7.54 & 22.09$\pm$9.38 \\
    NSMVC & \textbf{57.96$\pm$5.82} & \textbf{71.13$\pm$3.20} & \textbf{36.20$\pm$3.79} \\
    \hline
    \end{tabular}}%
  \label{6}%
\end{table}%

\begin{table}[!thb] 
  \normalsize
  \centering
  \caption{Results on Wisconsin.}
     \setlength{\tabcolsep}{3mm}{\begin{tabular}{cccl}
    \hline Methods  & ACC(\%) & Purity(\%) & NMI(\%) \\
    \hline
    KM(1) & 46.44$\pm$2.18 &48.67$\pm$1.82 & 5.69$\pm$2.30  \\
    KM(2) & \textbf{59.81$\pm$7.92} & 62.60$\pm$8.70 & 28.97$\pm$12.41  \\
    KM(All) & \textbf{58.57$\pm$6.93} & 60.33$\pm$7.68 & 25.95$\pm$11.35 \\
    Co-train & 42.58$\pm$1.89 & 52.57$\pm$1.10 & 8.28$\pm$0.83 \\
    Co-reg & 47.35$\pm$0.24 & 47.76$\pm$0.21 & 4.06$\pm$0.37 \\
    MVKKM & 45.62$\pm$2.88 & 48.03$\pm$1.34 & 6.29$\pm$2.22  \\
    AMGL & 47.09$\pm$0.16 & 47.55$\pm$0.00 & 4.03$\pm$0.31 \\
    CAMVC & 56.49$\pm$7.30 & 59.58$\pm$7.88 & 21.57$\pm$9.36 \\
    MSPL & 55.50$\pm$6.85 & 57.66$\pm$6.79 & 21.87$\pm$9.27 \\
    SAMVC & 49.93$\pm$3.13 & 48.93$\pm$3.15 & 7.04$\pm$4.69 \\
    DMVC  & 53.01$\pm$6.93 & 58.73$\pm$7.24 & 19.68$\pm$10.17 \\
    NSMVC & \textbf{60.30$\pm$4.89} & \textbf{73.70$\pm$1.73} & \textbf{40.48$\pm$1.85}\\
    \hline
    \end{tabular}}%
  \label{7}%
\end{table}%

\section{Experimental Results}
\subsection{Experimental Setup}\label{sec:exp_set}

\textbf{Data sets: } 
Handwritten Numerals\footnote{https://archive.ics.uci.edu/ml/datasets.php} data set is chosen from UCI machine learning repository, which contains 2000 points in 10 classes corresponded to numerals (0-9). 
Each instance is described by the following six views: 216 profile correlations, 76 Fourier coefficients of the character shapes, 64 Karhunen-Love coefficients, 6 morphological features, 240 pixel averages in 2 $\times$ 3 windows, and 47 Zernike moments.

MSRCv1\footnote{https://www.microsoft.com/en-us/research/project/image-understandin} is an image data set that constituted with 210 images over seven classes including tree, building, airplane, cow, face, car, and bicycle, each providing 30 images. 
For each image, it is described from 5 aspects: 24 Color Moments, 576 HOG features, 512 features, 256 LBP features, and 254 Centrist features.

The rest four data sets originate from the universities in Cornell, Texas, Washington and Wisconsin\footnote{http://www.cs.cmu.edu/afs/cs.cmu.edu/project/theo-20/www/data/}.
For each data set, there are two views, i.e., the content view and the citation view. 
According to the ground truth, these samples can be grouped into five classes: student, project, course, staff, and faculty. 

The detailed characteristics of data sets is shown in Table \ref{1}.
The image samples of Handwritten Numerals and MSRCv1 data sets are presented in Figure \ref{imgsamples}.

\textbf{Comparing Methods:}
To demonstrate the effectiveness of the proposed NSMVC model, we compare it with eight existing state-of-the-art multi-view clustering methods:
\begin{itemize}
\item Co-train: A co-training based approach for multi-view spectral clustering \cite{Kumar2011ACA}.
\item Co-reg: A centroid based co-regularized multi-view spectral clustering method \cite{Kumar:2011:CMS:2986459.2986617}.
\item MVKKM: The multi-view kernel $k$-means clustering method proposed by \cite{Tzortzis2012KernelBasedWM}.
\item AMGL: An auto-weighted multiple graph learning method for multi-view clustering \cite{AMGLNie}.
\item CAMVC: The robust capped-norm multi-view clustering method proposed by \cite{HUANG2018197}.
\item MSPL: A multi-view self-paced learning method for multi-view clustering \cite{Xu2015Multi}.
\item SAMVC: A self-paced and auto-weighted multi-view clustering method \cite{REN2020248}.
\item DMVC: An auto-weighted multi-view clustering method based on deep matrix decomposition \cite{HSDDMVC}.
\end{itemize}

In the above algorithms, MVKKM and CAMVC adopt the conventional linear combination strategy, AMGL, SAMVC, and DMVC apply the auto-weighting method for each view, and MSPL and SAMVC incorporate the self-paced learning strategy into multi-view clustering regimes to improve clustering results.

To make a comprehensive comparison, we also employ $k$-means clustering on each single view (e.g., KM(1) means applying KM on the first view).
Moreover, we concatenate the features of each view, and use $k$-means clustering on the joint view representation of the data, denoted as KM(All). 

\textbf{Implementation and Evaluation Metrics:}
For the KM method, we use the kmeans function in Matlab to form the clusters. For the MSPL, we follow the original paper to reproduce it. For the rest, we directly use the source codes from the authors and follow the suggesting parameter settings by corresponding publications. For all the methods, the number of clusters is always set to the number of ground truth classes. 
For the proposed method, the hyperparameters $\alpha$ and $T$ are selected in the ranges of \{0.3, 0.4, 0.5, 0.6, 0.7, 0.8\} and \{3, 4, 5, 6, 7, 8\}, respectively. 

In order to measure the quality of clustering results, we adopt three widely used evaluation metrics: accuracy (ACC), Purity, and normalized mutual information (NMI).
The higher value of each metric indicates the better performance. Each experiment was repeated for 30 times, and the mean and standard deviation of each metric in each data set were reported.


\begin{figure*}[!t]
\centering
\begin{minipage}{0.26\linewidth}
\centering
\includegraphics[width=1.9in]{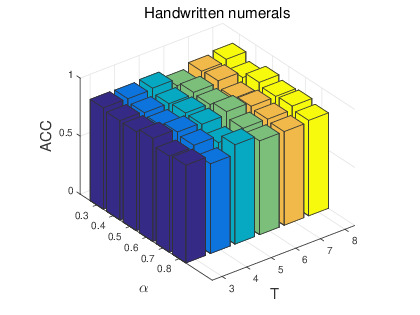}
\end{minipage}%
\begin{minipage}{0.26\linewidth}
\centering
\includegraphics[width=1.9in]{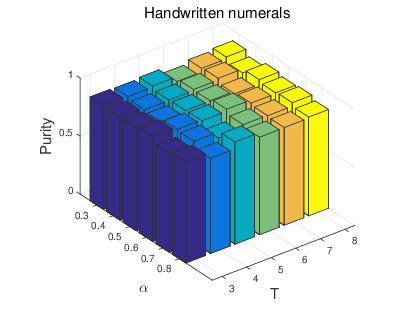}
\end{minipage}
\begin{minipage}{0.26\linewidth}
\centering
\includegraphics[width=1.9in]{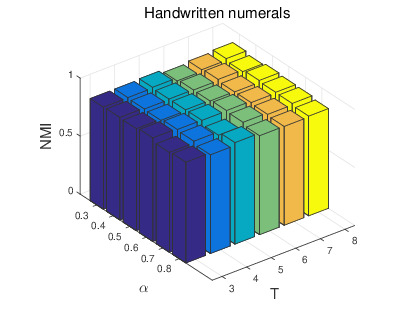}
\end{minipage}%
\caption{Clustering performance w.r.t. different parameter settings on Handwritten Numerals.}
\label{parameterhw}
\end{figure*}

\begin{figure*}[!t]
\centering
\begin{minipage}{0.26\linewidth}
\centering
\includegraphics[width=1.9in]{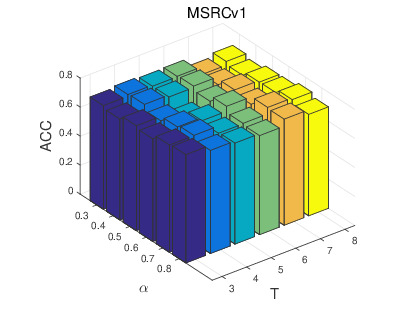}
\end{minipage}%
\begin{minipage}{0.26\linewidth}
\centering
\includegraphics[width=1.9in]{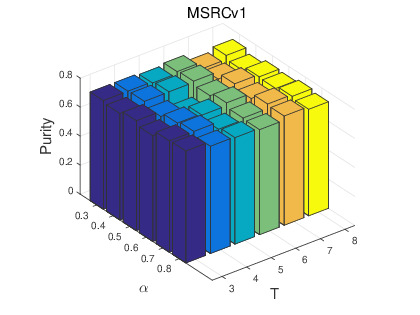}
\end{minipage}
\begin{minipage}{0.26\linewidth}
\centering
\includegraphics[width=1.9in]{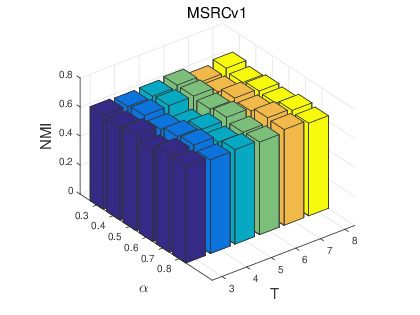}
\end{minipage}%
\caption{Clustering performance w.r.t. different parameter settings on on MSRCv1.}
\label{parameterMSRC}
\end{figure*}

\begin{figure*}[!t]
\centering
\begin{minipage}{0.26\linewidth}
\centering
\includegraphics[width=1.6in]{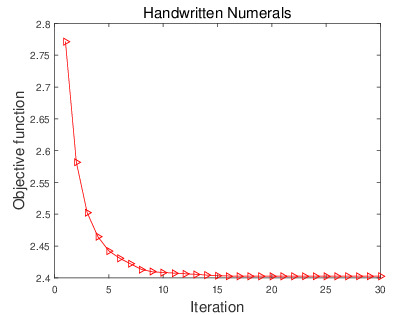}
\end{minipage}%
\begin{minipage}{0.26\linewidth}
\centering
\includegraphics[width=1.6in]{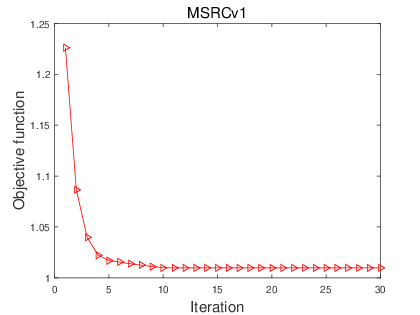}
\end{minipage}
\begin{minipage}{0.26\linewidth}
\centering
\includegraphics[width=1.6in]{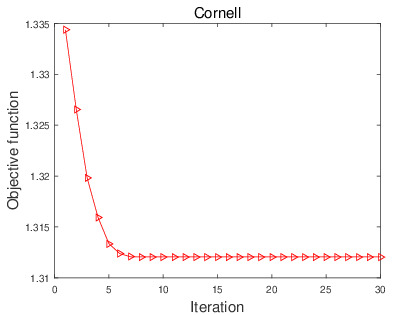}
\end{minipage}%

\centering
\begin{minipage}{0.26\linewidth}
\centering
\includegraphics[width=1.6in]{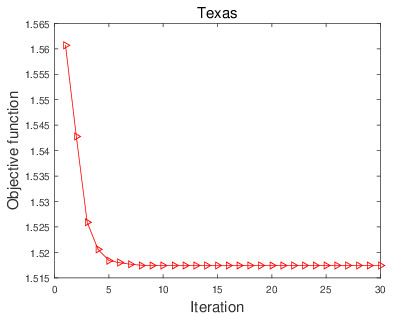}
\end{minipage}
\begin{minipage}{0.26\linewidth}
\centering
\includegraphics[width=1.6in]{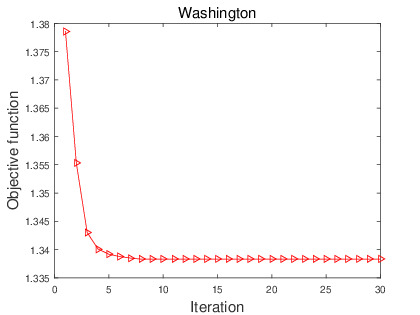}
\end{minipage}
\begin{minipage}{0.26\linewidth}
\centering
\includegraphics[width=1.6in]{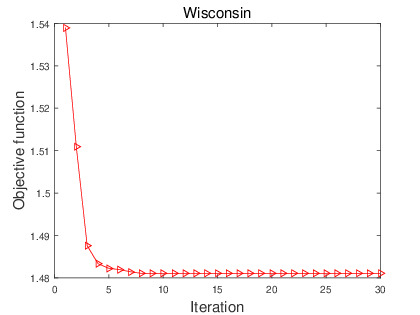}
\end{minipage}
\caption{Convergence curves of NSMVC on all data sets.}
\label{curve}
\end{figure*}

\subsection{Clustering Results}
\label{sec:res_real}
Tables \ref{2}-\ref{7} show the clustering performance of all the comparison methods on each data set in terms of ACC, Purity, and NMI. In each column, the best results and the comparable results under the $t$-test with $5\%$ significance level are highlighted in boldface. From these results, we have the following observations. 
First, the proposed NSMVC method almost always outperform the baseline methods on all data sets. 
This is mainly because our method adaptively assigns different exponents to different views according to their qualities.
Therefore, the proposed NSMVC reduces the negative impact of the corrupt views as well as maintains the influences of the more reliable views.
Moreover, it can be found that NSMVC always perform better than the conventional auto-weighted MVC methods like AMGL, DMVC, and SAMVC in which the exponents are consistent among all the views, which empirically confirm the effectiveness of the novel non-linear fusion technique. 
Further, by taking advantage of the self-paced learning to alleviate the non-convex issue, the proposed NSMVC also performs smaller standard deviations comparing with other methods, which indicates the better robustness of our method.

\subsection{Convergence Study}
\label{sec:convergence}
This section analyses the convergence of our method.
Figure \ref{curve} shows the convergence curves on six data sets in the first SPL process.
In these curves, the abscissa means the iteration number while the ordinate denotes the objective value of Eq. (\ref{eq:obj}).
From this figure, we can see that the proposed optimization algorithm converges quickly in the vicinity of the minimum, i.e., only around 15 iterations. 
At the same time, it is clear that even in the beginning of the training process, NSMVC converges quickly. As the training forwards, NSMVC learns more available knowledge for clustering and converges faster.


\subsection{Parameter Sensitivity}
In our NSMVC, there are two parameters for self-paced learning in Eq.~(\ref{beta}), i.e., the starting point $\alpha$ and the total iteration number $T$. 
Taking Handwritten Numerals and MSRCv1 data sets as examples, we examine the influence of these parameters to the clustering performance. Figures \ref{parameterhw} and \ref{parameterMSRC} show the variation of ACC, Purity and NMI over different $\alpha$ and $T$ on two data sets. We can observe that the clustering performance of the proposed NSMVC is relatively stable in a wide range of $\alpha$ and $T$ values, which may provide a good guidance for the parameter setting.

\section{Conclusion}
In this paper, a self-paced learning-based non-linear fusion method (NSMVC) is proposed to enhance the clustering performance of conventional multi-view clustering methods in both view level and instance level.
In the view level, a novel and effective non-linear fusion paradigm for multi-view clustering is proposed to exploit the complementary information from different views.
Meanwhile, to fit the non-linear model, we further design a novel modality of self-paced learning without the regularizer term.
By training the MVC model from simplicity to complexity progressively in the instance level, the non-convex issue is significantly alleviated and the robustness of the MVC model is further enhanced.
Extensive experiments on various multi-view data sets demonstrate the effectiveness of the proposed NSMVC.


\balance
\bibliographystyle{ACM-Reference-Format}
\bibliography{sample-base}

\end{document}